\documentclass[journal]{IEEEtran}
\usepackage{amsmath,amsfonts,algorithmic,algorithm,array}
\usepackage[caption=false,font=normalsize,labelfont=sf,textfont=sf]{subfig}
\usepackage{textcomp,stfloats,url,verbatim,graphicx,cite,booktabs,float}
\usepackage{amsmath,epsfig,multirow,graphicx,adjustbox,comment,lipsum}
\usepackage{hyperref}
\begin{document}

\title{DF4LCZ: A SAM-Empowered Data Fusion Framework for Scene-Level Local Climate Zone Classification}

\author{Qianqian Wu, Xianping Ma, Jialu Sui, and Man-On Pun
	\thanks{This work was supported by the National Key Research and Development Program of China under Grant No. 2020YFB1807700. \textit{(Corresponding author: Man-On Pun)}}
	\thanks{Q. Wu, X. Ma and J. Sui are with the Future Network of Intelligence Institute (FNii), the Chinese University of Hong Kong, Shenzhen, China (e-mails: qianqianwu@link.cuhk.edu.cn; xianpingma@link.cuhk.edu.cn; jialusui@link.cuhk.edu.cn).}
	\thanks{Man-On Pun is with the School of Science and Engineering, the Chinese University of Hong Kong, Shenzhen, China (e-mail: SimonPun@cuhk.edu.cn).}
}
\maketitle

\begin{abstract}
Recent advancements in remote sensing (RS) technologies have shown their potential in accurately classifying local climate zones (LCZs). However, traditional scene-level methods using convolutional neural networks (CNNs) often struggle to integrate prior knowledge of ground objects effectively. Moreover, commonly utilized data sources like Sentinel-2 encounter difficulties in capturing detailed ground object information. To tackle these challenges, we propose a data fusion method that integrates ground object priors extracted from high-resolution Google imagery with Sentinel-2 multispectral imagery. The proposed method introduces a novel Dual-stream Fusion framework for LCZ classification (DF4LCZ), integrating instance-based location features from Google imagery with the scene-level spatial-spectral features extracted from Sentinel-2 imagery. The framework incorporates a Graph Convolutional Network (GCN) module empowered by the Segment Anything Model (SAM) to enhance feature extraction from Google imagery. Simultaneously, the framework employs a 3D-CNN architecture to learn the spectral-spatial features of Sentinel-2 imagery. Experiments are conducted on a multi-source remote sensing image dataset specifically designed for LCZ classification, validating the effectiveness of the proposed DF4LCZ. The related code and dataset are available at \href{https://github.com/ctrlovefly/DF4LCZ}{https://github.com/ctrlovefly/DF4LCZ}.
\end{abstract}


\section{Introduction}
\begin{figure*}[htb]
	\centering
	\includegraphics[width=0.95\linewidth]{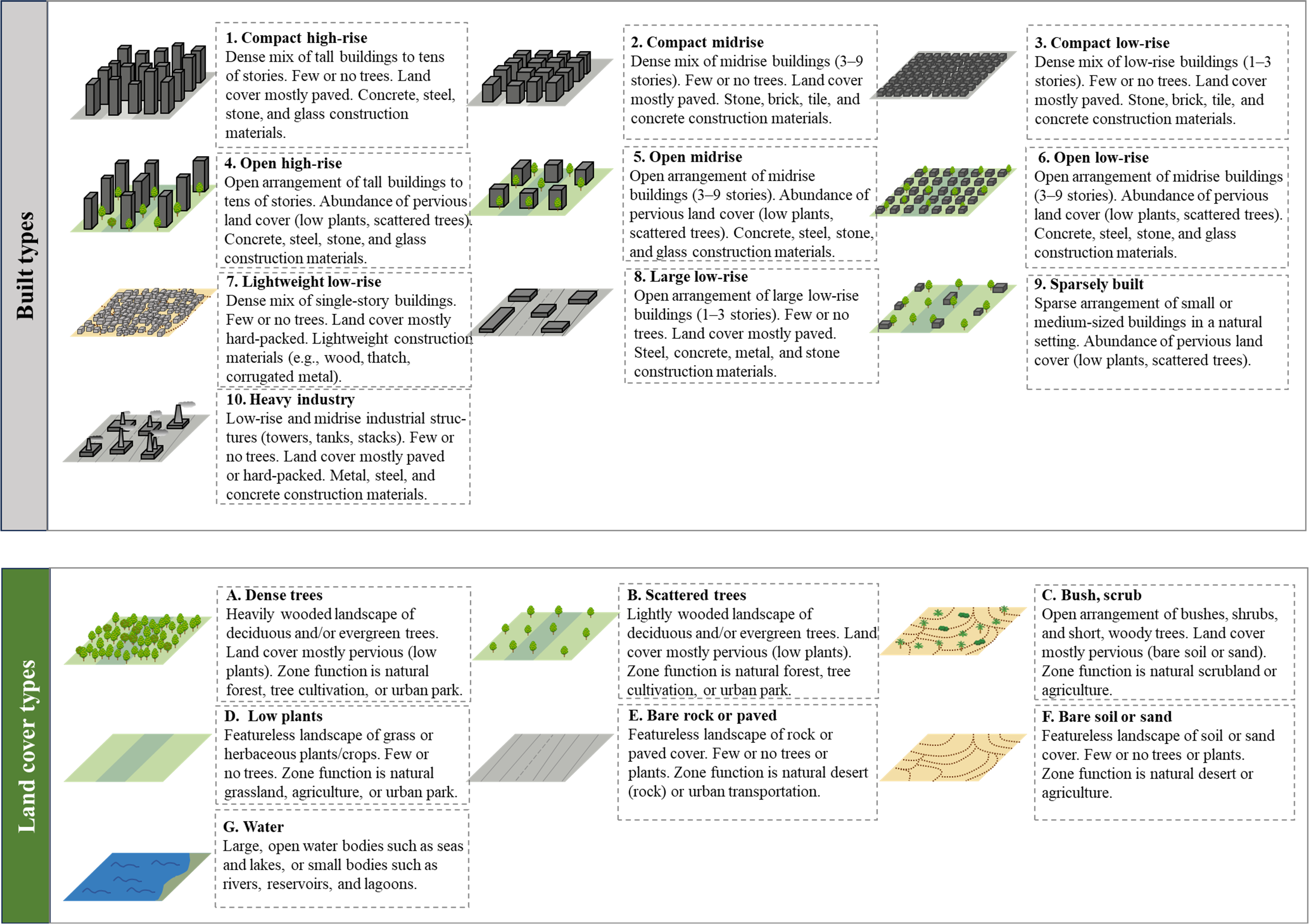}
	\caption{Illustration and a brief definition of LCZ types. There are ten built types (1–10) and seven land cover types (A–G).}
	\label{LCZ}
\end{figure*}

\IEEEPARstart{A}{} universal classification of urban regions is necessary for climatologists to facilitate comparisons of spatial variations in climatological data and communicate more effectively~\cite{stewart2014evaluation}. Acknowledging the inadequacies of traditional urban-rural divisions in describing the complexity and variety of cities, Stewart et al.~\cite{stewart2012local} developed a novel classification system for urban areas known as Local Climate Zones (LCZs) for Urban Heat Island (UHI) research. The system introduces $17$ distinct local climate zone categories comprised of $10$ built types and seven land cover types. Fig.~\ref{LCZ} shows the $17$ types of LCZs. According to the definition, each class describes areas with uniform landscape characteristics and human activity, typically extending horizontally from hundreds of meters to several kilometers.

LCZs provide a standardized framework for analyzing the thermal characteristics of different urban zones, thereby improving comparability across studies on urban heat islands~\cite{LECONTE201539,bechtel2019suhi}. However, achieving reliable and objective LCZ classification poses challenges, prompting extensive research efforts. Generally speaking, existing LCZ classification methods can be categorized into three groups, namely the RS-based, the Geographic Information System (GIS)-based, and combined methods~\cite{huang2023mapping}. Among the RS-based methods, scene-level LCZ classification emerges as a promising approach, assigning specific LCZ classes to image patches instead of individual pixels or objects. Recent research has shown that LCZ mapping should be approached as a scene classification task to comprehensively capture contextual information, as LCZ areas typically encompass the coexistence of multiple ground objects~\cite{liu2020local}. Thus inspired, this work aims to improve the classification accuracy of scene-level LCZ classification methods.

Deep learning models have been widely used for scene-level LCZ classification. In particular, CNN-based models like ResNet~\cite{chunping2018urban} and MSMLA-Net~\cite{kim2021local} have shown their effectiveness in scene-level classification. However, these CNN-based models suffer from a lack of explicit guidance in leveraging spatial location and arrangement information from ground objects. Unfortunately, such information has been proven essential for large-scale scene classification~\cite{peng2021instance, liang2020deep}.

Furthermore, another major challenge in scene-level LCZ classification resides in the inadequate spatial resolution of satellite imagery. For instance, Sentinel-2 imagery from the Copernicus Programme is frequently used for LCZ classification because it can capture multispectral data and is freely accessible. However, while LCZs concern complex local environments comprising various spatial objects, the Sentinel-2 imagery lacks sufficient spatial resolution to distinguish the ground objects. More specifically, the spatial resolution of Sentinel-2 imagery is often on the order of $10$ meters, making it challenging to discern the detailed visual information of objects. Moreover, the inherent blurriness impedes extracting distinctive features and identifying geographical connections among ground objects. While various studies have explored augmenting datasets with information from sources like Synthetic aperture radar (SAR) and OpenStreetMap (OSM) to enhance feature richness, the potential of merging high-resolution imagery, such as that from Google Earth, remains underexplored. High-resolution images such as Google Earth imagery with sub-meter spatial resolution offer distinct advantages, including more defined boundary and object details, significantly enriching the data derived from Sentinel-2 imagery by providing a wealth of supplementary information.

This study introduces an effective data fusion framework to address the mentioned challenges by incorporating high-resolution Google RGB imagery and Sentinel-2 imagery. To the best of our knowledge, this is the first study to leverage the complementary information between Google Earth and Sentinel-2 imagery for LCZ classification. Specifically, the proposed framework, named DF4LCZ, comprises dual streams aimed at utilizing instance-based location features from Google Earth imagery and scene-level spatial-spectral features from Sentinel-2 imagery. One stream employs a Segment Anything Model (SAM)-assisted Graph Convolutional Network (GCN) model to efficiently extract features from Google imagery, while the other utilizes a 3D ResNet11 model to fully extract features from Sentinel-2 multispectral imagery. Finally, a weighted fusion method combines the outputs of the two streams. Additionally, a new multi-source remote sensing image dataset is developed to evaluate the performance of DF4LCZ, comprising both Sentinel-2 multispectral imagery and their corresponding Google Earth RGB images.

The main contributions of this study are outlined as follows:
\begin{itemize}
	\item Given the limitations of Sentinel-2 satellite imagery, this study proposes utilizing high-resolution Google Earth imagery to enhance LCZ classification performance. To the best of our knowledge, this represents the initial attempt to integrate Google Earth RGB imagery and Sentinel-2 imagery for local climate zone classification;
	\item A dual-stream fusion framework, termed DF4LCZ, is developed to integrate instance-based location features with scene-level spatial-spectral features, deriving synergistic benefits from their complementary nature;
	\item DF4LCZ utilizes a GCN-based backbone for extracting scene-discriminative features from ground instances and exploits SAM to extract ground instances from Google Earth RGB images. Additionally, a 3D ResNet11 module is developed for extracting spatial-spectral features from Sentinel-2 imagery;
	\item Finally, a multi-source remote sensing image dataset, LCZC-GES2 (LCZ classification with Google Earth and Sentinel-2 imagery), is constructed utilizing Sentinel-2 multispectral and Google RGB images. Subsequently, extensive experiments are performed on this multi-source dataset to assess the performance of the proposed DF4LCZ method.
\end{itemize}

The rest of this paper is organized as follows. Section~\ref{sec:relatedwork} provides a review of related literature, followed by an elaboration on the proposed methodology in Section~\ref{sec:methodology}. Section~\ref{sec:dataset} introduces a new LCZ dataset to assess the performance of the proposed framework. After that, Section~\ref{sec:experiment} details the experimental setup, and Section~\ref{sec:results} assesses the performance of the proposed fusion technique. Finally, Section~\ref{sec:conclusion} provides the concluding remarks.	

\section{Related work}\label{sec:relatedwork}
\subsection{Scene-level LCZ classification}
In LCZ classification, RS-based methods are commonly categorized into three approaches: pixel-level, object-level, and scene-level classification. These approaches differ in the units of classification they employ. Specifically, the pixel-level approach assigns specific LCZ classes to individual pixels, whereas the object-level approach assigns LCZs to meaningful semantic objects. In sharp contrast, the scene-level approach labels each scene image {\em patch} with a single LCZ class. Due to the complexity of LCZ landscapes, the scene-level approach utilizing input images of larger sizes can effectively capture urban environmental features, resulting in more accurate performance~\cite{liu2020local}. Consequently, recent research advocates treating LCZ mapping as a scene classification task to comprehensively capture contextual information~\cite{yao2022mapping}. 

Deep learning models have been extensively utilized for scene-level classification. In particular, CNNs, are commonly employed in this context due to their effectiveness in image recognition~\cite{yoo2019comparison, rosentreter2020towards, feng2019embranchment, leichter2018improved, yang2019msppf, feng2020dynamic, yoo2020improving,zhao2020mapping, huang2021mapping}. One notable structure is the ResNet model. For example, Qiu et al.~\cite{qiu2018feature} utilized a ResNet as the classifier for LCZ classification in a large-scale study area covering nine cities in central Europe. Subsequently, various researchers have utilized ResNet and its derivatives for LCZ classification~\cite{jing2019effective, qiu2019local, kim2021local}. In particular, Qiu et al.~\cite{qiu2020multilevel} compared the performance of different ResNet architectures using the So2Sat dataset. These successful applications of ResNet models demonstrate their effectiveness in scene-level classification tasks. Additionally, some studies integrated channel attention mechanisms, such as squeeze-and-excitation blocks~\cite{liu2020local, zhou2022deep} and convolutional block attention modules~\cite{wang2023llnet}, into ResNet backbones, improving the model's capability to emphasize essential features and enhance accuracy.  

Nonetheless, there are inherent limitations associated with CNNs in scene-level LCZ classification. For instance, current CNN methodologies lack explicit guidelines for utilizing spatial location and arrangement information from ground objects, which have been demonstrated to be crucial cues for scene classification~\cite{liang2020deep, shen2022graph}. In this study, we propose the incorporation of ground instance priors to enhance LCZ classification accuracy. Furthermore, while most existing studies utilize 2D CNNs for spatial information extraction, multispectral satellite images contain both spatial and spectral information. Therefore, we employ a 3D ResNet model to extract spatial and spectral information from Sentinel multispectral imagery jointly.

\subsection{Representation of object features}
Given that LCZs can be regarded as complex local scenarios comprising diverse objects leveraging component object information can significantly improve the performance of scene recognition models. In particular, investigating the geographical connections among objects aids in mitigating the challenges arising from high intra-class variability and low inter-class variability among different LCZ types. Several studies have designed methods to extract local objects to achieve discriminative scene representation. Wang et al.~\cite{wang2016modality} used a publicly available proposal extractor to identify critical components contributing to scene discriminability, whereas Chen et al.~\cite{chen2022remote} studied semantic areas using CNN-based heat maps in the corresponding original image, thus enhancing scene representations. However, these studies overlook spatial connection information among spatial objects.

In recent years, Graph Neural Networks (GNNs) have emerged as a revolutionary advancement in machine learning, exhibiting exceptional performance in tasks involving structured data, such as social networks, molecules, and recommendation systems~\cite{zhou2020graph}. Their unique architecture, which iteratively aggregates information from neighboring graph nodes, sets them apart from other models. Recently, GNNs have been commonly employed for extracting the spatial features of component objects in RS and medical images. Liang et al. \cite{liang2020deep} conducted the initial investigation into the dependencies among ground objects in remote sensing scene classiﬁcation tasks. They detected ground objects using the Faster R-CNN detector, constructed them into a graph, and utilized a GCN model to learn spatial location features from the graph. Furthermore, Shen et al. \cite{shen2022graph} employed CA2.5-Net as the backbone for nuclei segmentation and integrated a GNN module to capture the geometric aspects of cell-level spatial information. Recently, Peng et al. \cite{peng2022multi} conducted simultaneous pixel-level prediction of component objects and recognition of facilities by employing a GNN to aggregate spatial relations.

While these methods have provided valuable insights into representing object location and arrangement features using GNNs, they are carried out for a single data source, limiting the information to that particular source. In this work, we propose fusing the instance-based location and arrangement information extracted from the Google Earth imagery with the spatial-spectral features derived from Sentinel-2 imagery, leveraging complementary multi-source data. Additionally, some studies require training task-specific networks to detect component objects, which is time-consuming and restricted to specific ground instance categories. Therefore, in this work, we propose employing a SAM~\cite{kirillov2023segment} model to extract ground instances, benefiting from its generalization ability to unseen data and avoiding training a task-specific object detection model. SAM is a foundation model designed specifically for image segmentation~\cite{kirillov2023segment}, trained on an extensive dataset comprising over one billion segmentation masks from $11$ million meticulously curated images. It distinguishes itself by enabling zero-shot generalization to previously unseen objects without the need for additional training data. Within the realm of RS, the SAM model has been employed in various specific tasks, particularly in remote sensing semantic segmentation tasks~\cite{ma2023sam, zhang2023text2seg, wang2024samrs}, showcasing its ability to enhance ground instance segmentation.

\begin{figure*}[htp]
	\centering
	\includegraphics[width=0.95\linewidth]{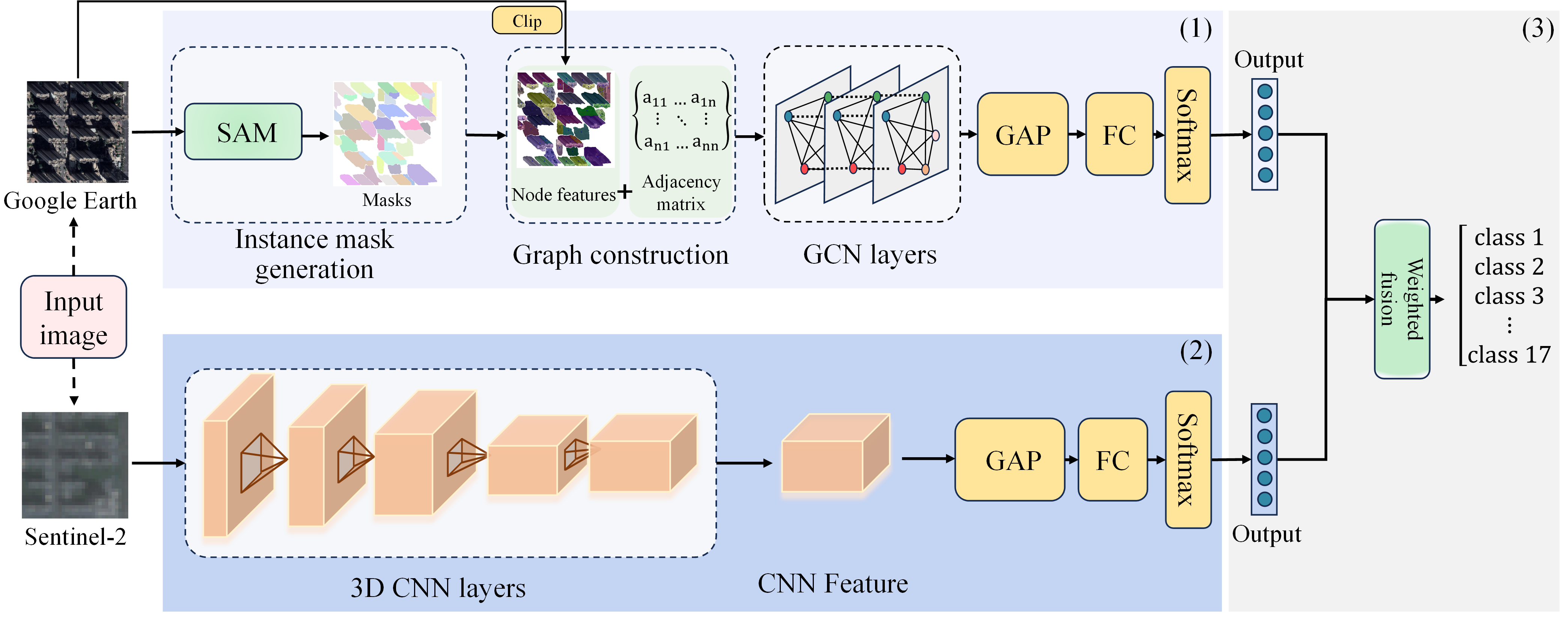}
	\caption{Illustration of the DF4LCZ framework, composed of three main modules: (1) a Google Earth stream focusing on instance-based location feature extraction; (2) a Sentinel-2 stream dedicated to scene-level spatial-spectral feature extraction; and (3) a fusion and classification module.}
	\label{fig:framework}
\end{figure*}

\section{Methodology}\label{sec:methodology}
The proposed dual-stream framework is depicted in Fig.~\ref{fig:framework}. As shown, the framework comprises three primary modules: (1) a Google Earth stream focusing on instance-based location feature extraction; (2) a Sentinel-2 stream dedicated to scene-level spatial-spectral feature extraction; and (3) a fusion and classification module. Further details regarding these modules are provided below.

\subsection{Instance-based location feature extraction}

\subsubsection{SAM-assisted instance extraction}
The SAM-assisted instance extraction module is designed to generate segmentation masks that delineate the boundaries of ground instances. SAM can avoid the computational cost of retraining a task-specific object detection network, which is advantageous. In this module, the SAM model generates segmentation masks and accompanying bounding box information by sampling single-point input prompts in a grid across the input image patch. Figure~\ref{fig:framework} (l) depicts an example of the segmentation masks. Ground instances from the input Google Earth image patch can be extracted using the output masks through clipping. Subsequently, these instances are regarded as valuable domain knowledge or priors for the next-stage processing, thereby enriching the information available for classifying the complex scenes of LCZs.

\subsubsection{Graph construction}
Next, we construct a graph using the extracted segmentation masks and accompanying bounding boxes. We first compute the center location $v=(c_{x}, c_{y})$ of each bounding box generated by SAM before treating it as a single node in the graph. Subsequently, the graph's adjacency matrix $\mathbf{A}$ is derived from the distances between each pair of center locations. After that, the Google Earth image patch is clipped with the masks generated by SAM to obtain the instance pixels. Using the mean pixel value $(r,g,b)$ of each instance and the center location $(c_{x}$, $c_{y})$, we can form the node feature of an instance as $\mathbf{x}_v=[r, g, b, c_{x}, c_{y}]$. Finally, we obtain the graph $G= (\mathbf{A},\mathbf{X})$ with $\mathbf{X}=[\mathbf{x}_{v_1} \mathbf{x}_{v_2} \cdots  \mathbf{x}_{v_n} ]^T$, representing the matrix of node features, and $\{v_{1},v_{2},\ldots,v_{n}\}$ representing the set of graph nodes.

\begin{figure*}[h]
	\centering
	\includegraphics[width=0.95\linewidth]{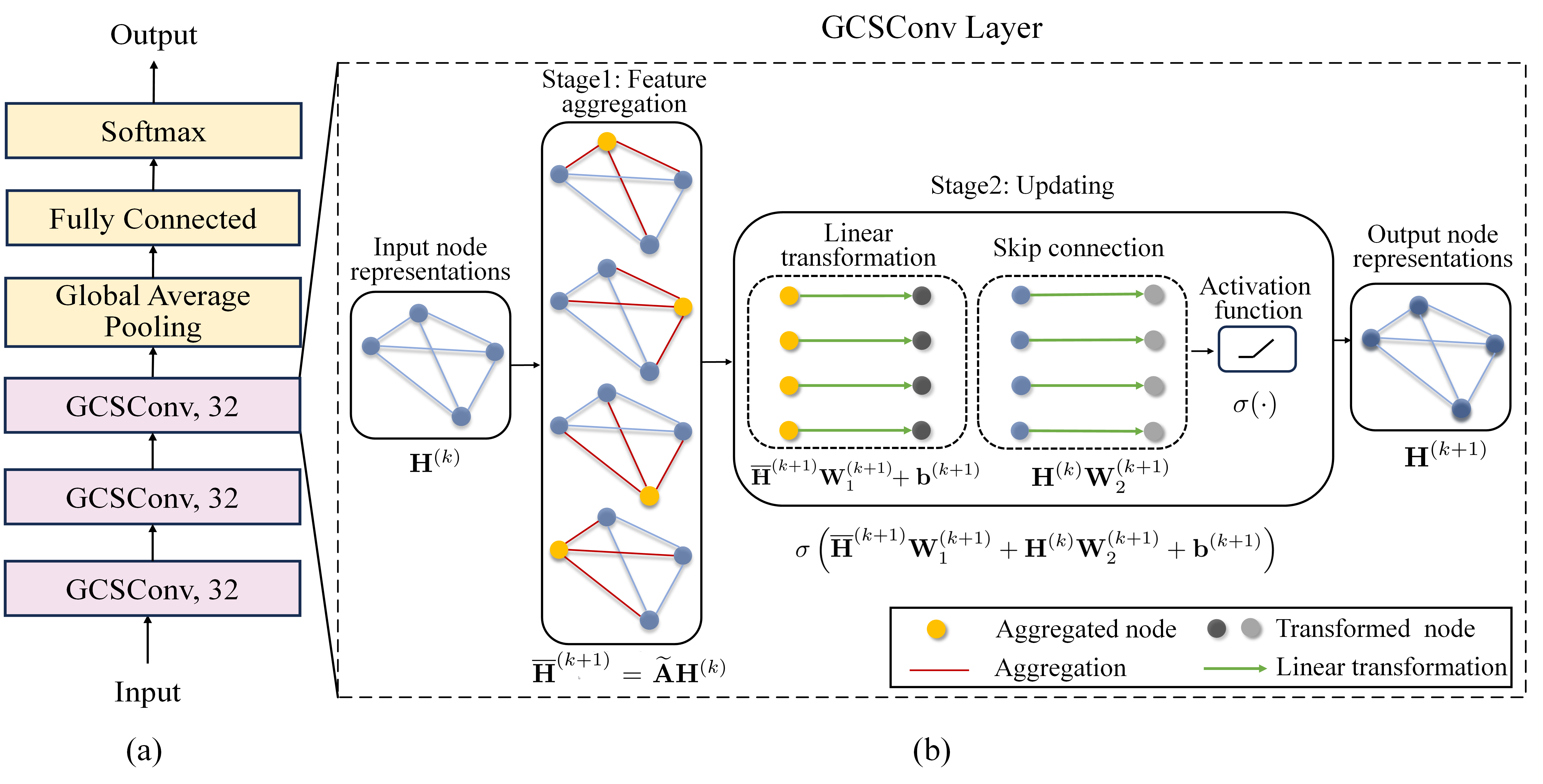}
	\caption{Structure of the GCN Network in this study. (a) Input, Layers, and Output: The GCN network's input is the graph obtained from the graph construction procedure. The network consists of three graph convolutional layers with skip connections (GCSConv), each producing an output dimension of 32 for every node embedding vector $h^{(k+1)}$, followed by a global average pooling layer, a fully connected layer, and a softmax layer. The output of the network is a vector representing probability distribution over all the possible classes. (b) GCSConv Layer: The GCSConv layer embeds input node features through two stages, namely feature aggregation and updating.}
	\label{fig_method1}
\end{figure*}

\subsubsection{Graph convolutional network}
Finally, the GCN network \cite{kipf2016semi} is utilized to capture spatial location and arrangement features from the instance-based graph. The GCN is chosen due to its proficiency in capturing complex spatial relationships inherent in graph structures. In this study, there are three graph convolutional layers with skip connections (GCSConv) for graph feature extraction, as depicted in Fig.~\ref{fig_method1} (a).

Each layer consists of two stages, namely feature aggregation and updating, as shown in Fig.~\ref{fig_method1} (b). The aggregation stage involves collecting information from neighboring nodes. Assuming that the three graph convolutional layers are indexed using $k \in \{0,1,2\}$, for the $(k+1)$-th graph convolutional layer, the aggregated node feature matrix $\overline{\mathbf{H}}^{(k+1)}$ is computed as $\overline{\mathbf{H}}^{(k+1)} = \widetilde{\mathbf{A}} \mathbf{H}^{(k)}$, where $\widetilde{\mathbf{A}}$ is the normalized $\mathbf{A}$ and given as 
\begin{equation}\label{eq1}
	\widetilde{\mathbf{A}}=\mathbf{D}^{-\frac{1}{2}} \mathbf{A} \mathbf{D}^{-\frac{1}{2}}.
\end{equation}
In Eq. (\ref{eq1}), $\mathbf{D}$ is the degree matrix of $\mathbf{A}$. Additionally, $\mathbf{H}^{(k)}$ represents the input node feature matrix of the layer, and the initial node feature matrix is $\mathbf{H}^{(0)} = \mathbf{X}$.

The updating stage refines the node representations by incorporating aggregated features and introduces a skip connection to preserve original features. We denote the output feature representation as $\mathbf{H}^{(k+1)}$. The updating rule of the $(k+1)$-th layer is given by

\begin{equation}
	\mathbf{H}^{(k+1)} = \sigma\left(\overline{\mathbf{H}}^{(k+1)} \mathbf{W}^{(k+1)}_{1} + \mathbf{H}^{(k)} \mathbf{W}^{(k+1)}_{2} + \mathbf{b}^{(k+1)}\right),
\end{equation}
where $\sigma(\cdot)$ denotes the activation function while $\mathbf{W}_1$, $\mathbf{W}_2$, and $\mathbf{b}$ are trainable parameters. 

After the graph convolutional layers, a global average pooling layer is employed to reduce the dimensionality of the output node features to a 1D vector. Subsequently, this vector is passed to a fully connected layer, and its output is then directed to a softmax layer for generating class probability estimates for the Google Earth stream.

\subsection{Scene-level spatial-spectral feature extraction}
\begin{figure*}[!t]
	\centering
	\includegraphics[width=0.75\linewidth]{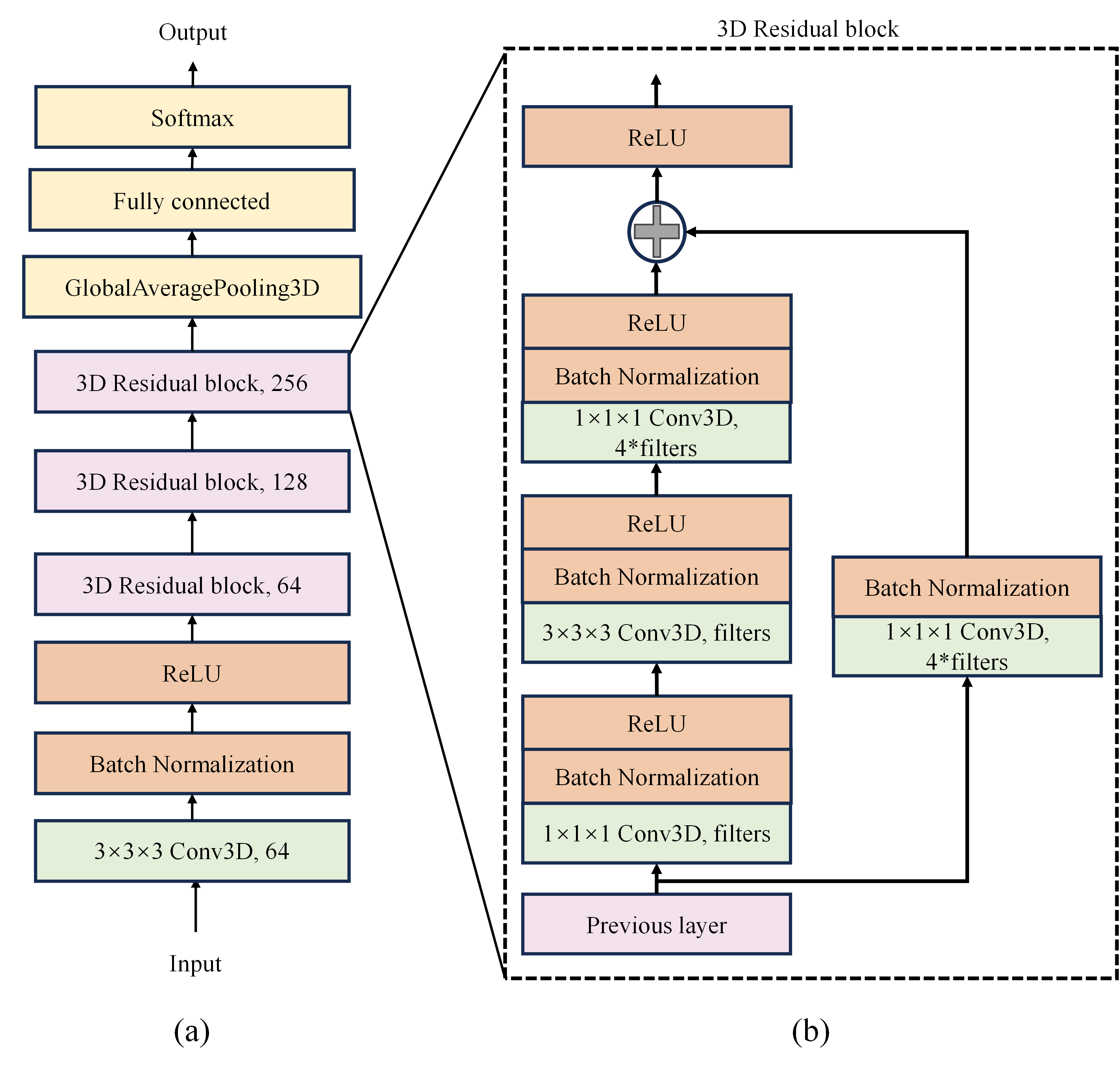}
	\caption{Structure of the 3D ResNet11 network in this study. (a) Input, Layers, and Output: The 3D ResNet11 network's input is the Sentinel-2 image patch. The network consists of an initial convolutional layer with an output channel dimension of 64, followed by a batch normalization layer and ReLU. Following that, three 3D residual blocks are applied to extract spatial-spectral features. The stacked residual blocks are followed by a global average pooling layer, a fully connected layer, and a softmax layer. The output of the network is a vector representing probability distribution over all the possible classes. (b) 3D residual block: Each 3D residual block has a sequence of 3D convolutional layers followed by batch normalization and rectified linear unit activation functions, and a residual connection is within each block. The number of filters for the three blocks is 64, 128, and 256, respectively.}
	\label{fig_method2}
\end{figure*}

In order to effectively leverage features from Sentinel-2 multispectral imagery, we employ a 3D residual network architecture referred to as 3D ResNet11. This network draws inspiration from the successful application of 2D ResNets in image recognition and extends this concept into three-dimensional spatial-spectral domains. The structure of the 3D ResNet11 network in this study is depicted in Fig.~\ref{fig_method2} (a).

We denote by $\textbf{I} \in \mathbb{R}^{M \times N \times D}$ the Sentinel-2 image patch, where $M$ and $N$ denote the patch width and height, respectively, while $D$ signifies the number of spectral bands. As the input to the 3D ResNet11 network, each $\textbf{I}$ is fed into the network, with the network's outputs being the probability estimates. The 3D ResNet11 network primarily consists of one initial 3D convolutional layer and three 3D residual blocks. 

In the 3D convolutional layer, the value located at position ($x, y, z$) within the $j$-th feature cube in the $i$-th layer can be expressed as
\begin{equation}
	v_{i j}^{x y z}=g\left(\sum_m \sum_{h=0}^{H_i-1} \sum_{w=0}^{W_i-1} \sum_{c=0}^{C_i-1} k_{i j m}^{h w c} v_{(i-1) m}^{(x+h)(y+w)(z+c)}+b_{i j}\right),
\end{equation}
where $g(\cdot)$ represents the activation function and $m$ denotes the index of the feature cube connected to the current feature cube in the ($i-1$)-th layer. The term ``feature cube" is analogous to ``feature map" in 2D CNNs \cite{yang2019multi}. Additionally, $k_{i j m}^{h w c}$ denotes the ($h,w,c$)-th value of the kernel associated with the $m$-th feature cube in the previous layer, and $b_{ij}$ represents the bias on the $j$-th feature cube in the $i$-th layer. Finally, the length and width of the convolution kernel in the spatial dimensions are represented by $H_{i-1}$ and $W_{i-1}$ respectively, while the kernel size in the spectral dimension is denoted by $C_{i-1}$.

The 3D convolutional layers are structured into three residual blocks. Each block comprises a sequence of 3D convolutional layers followed by batch normalization and rectified linear unit activation functions, as depicted in Fig.~\ref{fig_method2} (b). Within each block, residual connections help maintain the flow of information across the network to mitigate the vanishing gradient problem~\cite{he2016deep}. The batch normalization helps to prevent internal changes in covariance~\cite{ioffe2015batch}.

After the three residual blocks, a 3D global average pooling layer is employed to reduce the dimensionality of the output feature cubes to a 1D vector. Subsequently, this vector is fed into a fully connected layer, followed by a softmax layer, generating class probability estimates for the Sentinel-2 stream.

\subsection{Fusion and Classification}

In this study, we employ a decision-level fusion method known as weighted fusion~\cite{roitberg2022comparative} to link the information from both Google Earth images and Sentinel imagery for scene-level local climate zone classification. The weighted fusion combines the predictions of the Google Earth stream and the Sentinel-2 stream based on their class probability estimates $\mathbf{c}_g\in \mathbb{R}^{d}$ and $\mathbf{c}_s\in \mathbb{R}^{d}$, where $d$ denotes the number of classes. The weighted fusion can be mathematically formulated as follows:

\begin{equation}
	f(\mathbf{c}_u) = \alpha * \mathbf{c}_{g} + (1-\alpha) * \mathbf{c}_{s}, 
	\label{eq:fusion}
\end{equation}
where $f(\cdot)$ represents the weighted fusion whereas $\mathbf{c}_u$ denotes the set of all probability estimates. Furthermore, $\alpha \in \left[0,1\right]$ is a weighting coefficient.

\begin{figure}[!t]
	\centering
	\includegraphics[width=3.5in]{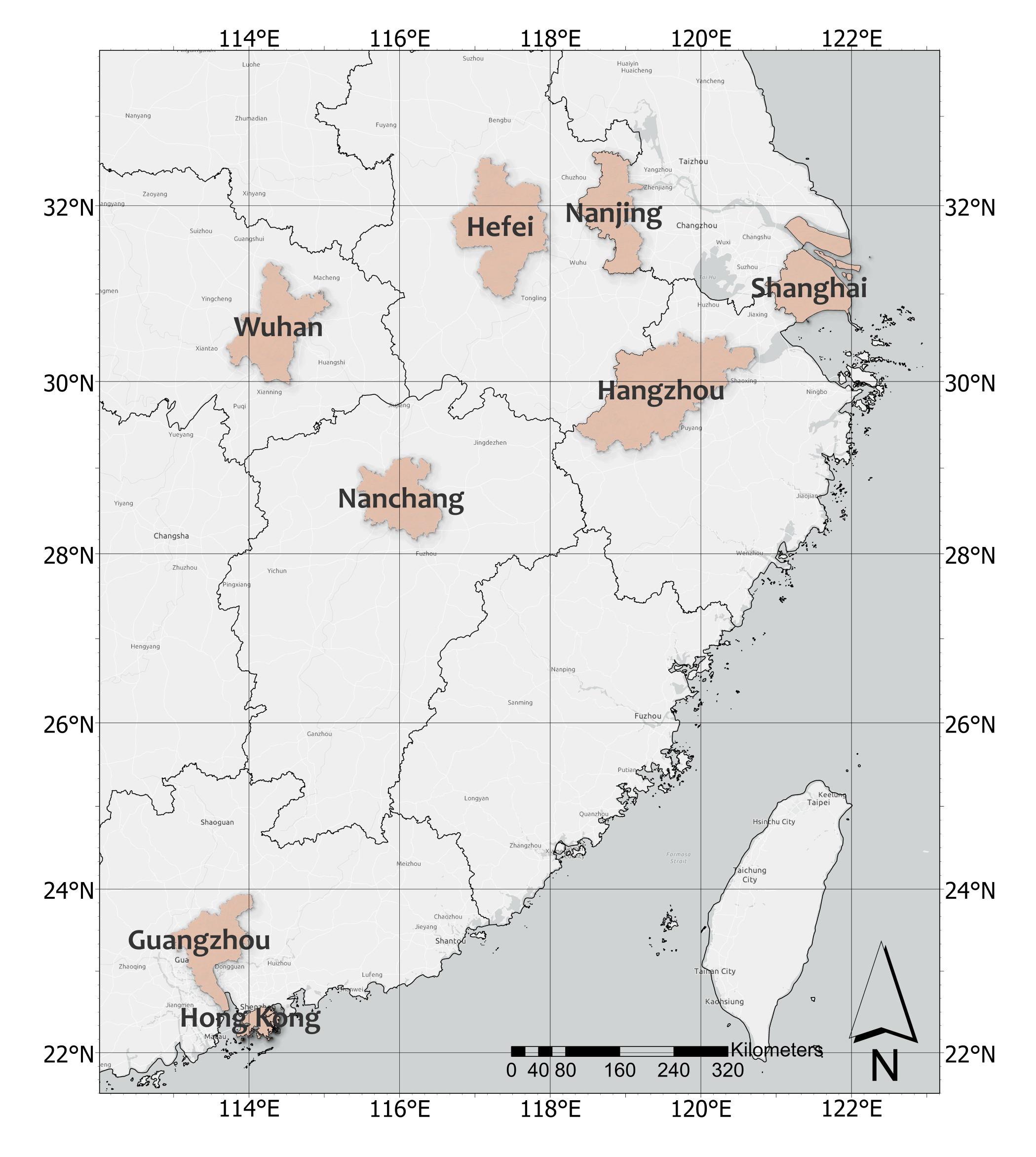}
	\caption{An illustration of our study area spanned eight cities in Southeast China, namely Guangzhou, Hangzhou, Hefei, Hong Kong, Nanchang, Nanjing, Shanghai, and Wuhan.}
	\label{fig_1}
\end{figure}

\section{The Proposed LCZC-GES2 Dataset}\label{sec:dataset}

LCZ datasets are crucial for training deep feature learning models. However, there is a lack of multi-source LCZ datasets containing both Google Earth and Sentinel-2 images for LCZ classification. Furthermore, adding additional image sources to existing datasets, such as So2Sat~\cite{zhu2020so2sat}, is challenging due to the absence of geographic coordinates used in generating these datasets. Therefore, to assess the performance of the proposed DF4LCZ, a dataset called LCZC-GES2 (LCZ classification with Google Earth and Sentinel-2 imagery) has been created using images from both Google Earth and Sentinel-2. Specifically, data were collected from eight cities in Southeast China: Guangzhou, Hangzhou, Hefei, Hong Kong, Nanchang, Nanjing, Shanghai, and Wuhan. Fig.~\ref{fig_1} displays the spatial distribution of the selected cities on a map. The selection of these cities is justified by their substantial population sizes, making them significant contributors to the urban landscape of Southeast China. Additionally, the selected cities exhibit varied geographical distributions, resulting in a diverse array of urban structures. These diverse urban structures provide a rich dataset for classifying local climate zones.

The LCZC-GES2 dataset comprises 19088 pairs of image patches. Each pair includes a Google Earth RGB image and a Sentinel-2 multispectral image patch. Table~\ref{tab:samples} shows the number of image patch pairs per LCZ class. The origin and acquisition process of the LCZC-GES2 labels and images are detailed as follows.

\begin{table}[h]
\centering
\caption{Numbers of image patch pairs per LCZ class. Each pair consists of a Google Earth RGB image and a Sentinel-2 multispectral image patch.}
\label{tab:samples}
\begin{tabular}{lclc}
	\hline Class & Number of pairs & Class & Number of pairs \\
	\hline LCZ1 & 1012 & LCZ10 & 736 \\
	LCZ2 & 1030 & LCZA & 1600 \\
	LCZ3 & 1692 & LCZB & 1108 \\
	LCZ4 & 1103 & LCZC & 604 \\
	LCZ5 & 903 & LCZD & 936 \\
	LCZ6 & 594 & LCZE & 987 \\
	LCZ7 & 22 & LCZF & 1116 \\
	LCZ8 & 2722 & LCZG & 2207 \\
	LCZ9 & 716 & & \\
	\hline
\end{tabular}
\end{table}

\begin{figure*}[t]
	\centering
	\includegraphics[width=\linewidth]{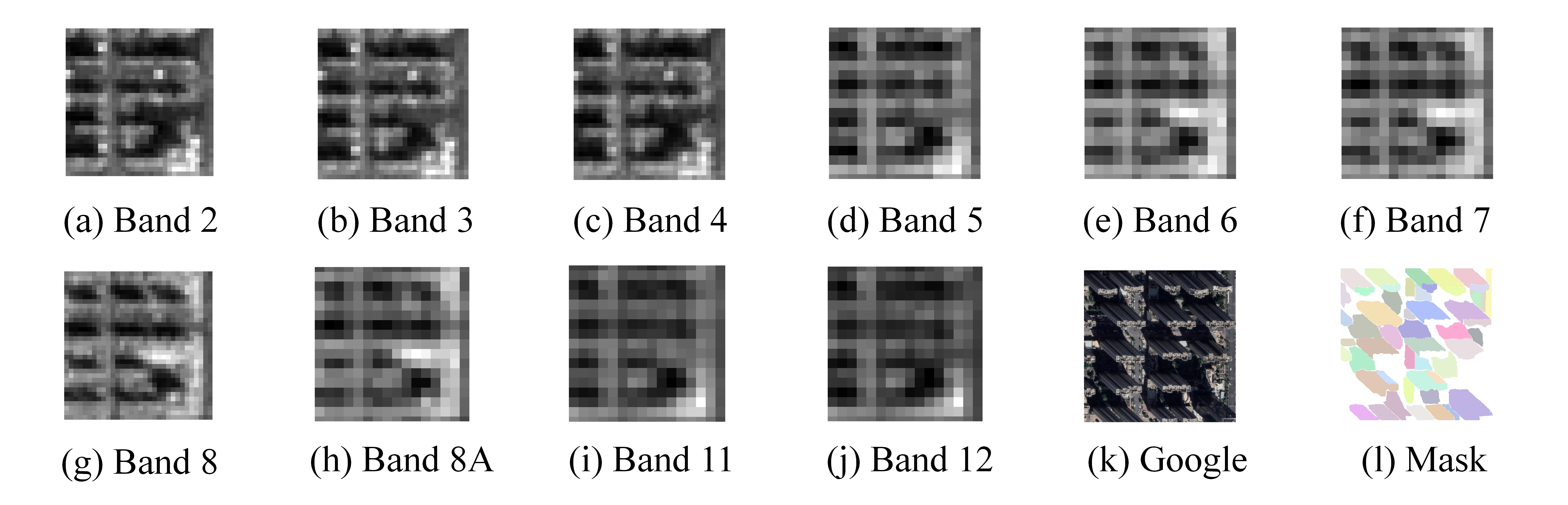}
	\caption{Illustration of the Sentinel-2 and Google Earth image patches. (a) - (j) are ten bands of the Sentinel-2 image patch, (k) is the Google Earth image patch, and (l) is the segmentation mask generated from the SAM model.}
	\label{fig:data1}
\end{figure*}

\begin{figure*}[t]
	\centering
	\includegraphics[width=\linewidth]{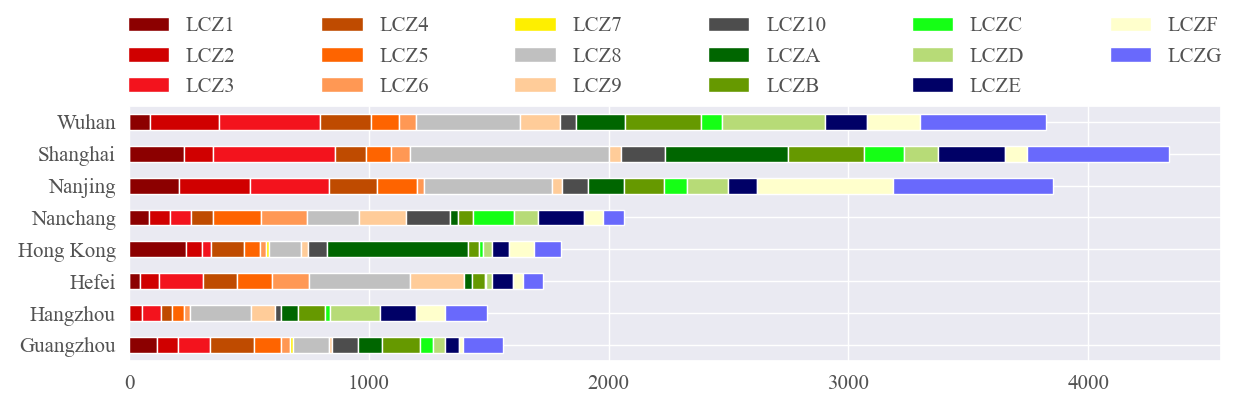}
	\caption{Number and distribution of labeled LCZ sample pairs of eight cities.}
	\label{fig:data2}
\end{figure*}

\subsubsection{LCZ Reference Data}
The dataset labels are sourced from the World Urban Database and Access Portal Tools (WUDAPT)~\cite{ching2018wudapt}. WUDAPT is a tool used to create local climate zone maps for individual cities, employing a standardized methodology based on a random forest (RF) classifier. WUDAPT obtains its dataset through crowdsourcing, enabling users to create training areas (TAs) that delineate LCZ types in different urban locations~\cite{demuzere2021lcz}. These TAs are depicted as digital polygons labeled with LCZ types. In our study, we acquired the TAs from WUDAPT to serve as the reference data for each study city. In order to mitigate errors and balance the sample distribution, we refined the acquired crowdsourced labels and introduced a few additional TAs.

\subsubsection{Sentinel-2 imagery}
Sentinel-2 imagery, acquired from the European Space Agency's (ESA) Sentinel-2 satellite constellation, includes a comprehensive range of 13 spectral bands~\cite{drusch2012sentinel}. These bands span the visible, near-infrared, and shortwave infrared spectra, each with a Ground Sample Distance (GSD) of $10$, $20$, or $60$ meters. This study selected Sentinel level-2A (L2A) data, which provides orthoimage atmospherically corrected surface reflectance products. The data range was filtered from January 1, 2023, to June 30, 2023, with a cloud pixel percentage of less than $3\%$.

Sentinel-2 imagery underwent preprocessing involving several steps. All ten bands out of the 13 were initially stacked to create a multispectral composite image. Fig.~\ref{fig:data1} (a)-(j) illustrates an example of the $10$ bands. Bands $1$, $9$, and $10$ were excluded from the analysis due to their minimal relevance to LCZ classification, aligning with current literature. After that, resampling was performed to standardize the spatial resolution to $10$ meters for all used bands. Afterward, the images' Digital Number (DN) values were normalized to a reflectance range of $0$ to $1$.

\subsubsection{Google Earth imagery}
The Google Earth images utilized in this study are sourced from Google Earth, a widely used program for viewing very high-resolution remote sensing imagery~\cite{lisle2006google}. The imagery is reportedly sourced from various providers, primarily commercial satellites and aerial photography sources. The imagery has a spatial resolution finer than $5$ meters and consists of RGB bands. In our work, QGIS~\cite{flenniken2020quantum}, a robust open-source GIS software, was utilized to access Google Earth imagery globally and acquire corresponding images of the study area. The raw images downloaded have a spatial resolution of $0.3$ meter. Subsequently, the imagery underwent resampling to achieve a spatial resolution of $1$ meter. Concurrently, the imagery was meticulously co-registered with the acquired Sentinel-2 imagery. Fig.~\ref{fig:data1} (k) shows a Google Earth image patch.

\subsubsection{Patch extraction and sampling strategy}\label{sec:sampling}
Image patches for scene-level classification were generated using a patch extraction method like the one described in~\cite{zhu2020so2sat}. Firstly, we sampled a specified number of points inside the reference polygons. These points serve as the centroids for extracting image patches from the input data. Image patches were sized at $32 \times 32$ pixels for Sentinel imagery and $320 \times 320$ pixels for Google imagery. Subsequently, the patches were meticulously refined by eliminating those with minimal overlap with the reference polygons, ensuring an accurate representation of regions with homogeneous LCZ characteristics. In total, $19,088$ pairs of images were compiled from Sentinel-2 and Google Earth imagery sources. The distribution of the patches is shown in Fig.~\ref{fig:data2}. Finally, the images generated by patch extraction were partitioned into training, testing, and validation sets for training the deep learning model, following a proportional distribution of $7:2:1$, respectively.

Two distinct sampling strategies were employed in the patch extraction process. One strategy, named ``splitting the sample pool", involves stratifying the dataset by dividing the sample points drawn from all TAs. Alternatively, the other strategy, named ``splitting the polygon pool", initially divides reference polygons to create training and test sets; subsequently, points are sampled within these polygons. We utilized the dataset generated from the former strategy for most experiments and the latter for evaluating transferability. A comprehensive discussion of these two strategies is provided in~\cite{xu2021application}. 

\section{Experiment Setup}\label{sec:experiment}
\subsection{Network training}
Essential implementation specifics are detailed below. Augmentation techniques, including horizontal and vertical flips, as well as rotations, are employed to expand the dataset and mitigate overfitting. A cross entropy-based loss function is employed while the adaptive moment estimation (Adam) optimizer is utilized to adjust learning rates during training dynamically. Additionally, the training parameters consist of a batch size of $64$, a maximum of $100$ epochs, and a learning rate of $0.002$ with a decay factor of $0.4$. Early stopping is implemented to halt training when performance plateaus, preventing overfitting and expediting convergence. 

Furthermore, the proposed dual-stream framework undergoes a two-phase process. Initially, each single-stream baseline is trained individually. Subsequently, the weighted fusion is applied with the parameters of each single-stream baseline being frozen. The weighting coefficient $\alpha$ is set to 0.6. 

Our simulation platform was developed using Python 3.8, with Keras version 2.11.0 and TensorFlow version 2.11.0 as the primary deep learning framework. Additionally, Spektral version 1.3.0~\cite{grattarola2021graph} was employed for constructing GCN, and SAM of version 1.0 was utilized for instance segmentation. Experiments were performed on a computational platform featuring an NVIDIA GeForce RTX $3090$ GPU with $24$ GB of memory, a 12th Gen Intel(R) Core(TM) i9-12900K CPU, and $126$~GB of RAM. 

\begin{table*}[htp]
	\centering
	\caption{Overall results of the DF4LCZ model and the single-stream baselines.}
	\label{tab:results}
	\begin{adjustbox}{width=0.5\textwidth}
		\begin{tabular}{lccccc}
			\hline
			Data & OA & OA$_{BU}$ & OA$_{N}$ & Kappa & Avg. F1 \\ \hline
			G & 51.93\% & 55.28\% & 47.81\% & 0.4775 & 50.66\% \\
			S & 92.42\% & 91.53\% & 93.50\% & 0.9179 & 92.42\% \\
			S+G & \textbf{93.81\%} & \textbf{93.10\%} & \textbf{94.68\%} & \textbf{0.9329} & \textbf{93.80\%} \\ \hline\\
		\end{tabular}
	\end{adjustbox}
	\footnotesize \par
	Abbreviations: G - Google Earth Image, S - Sentinel Image
\end{table*}

\begin{figure*}[htp]
	\centering
	\includegraphics[width=0.95\linewidth]{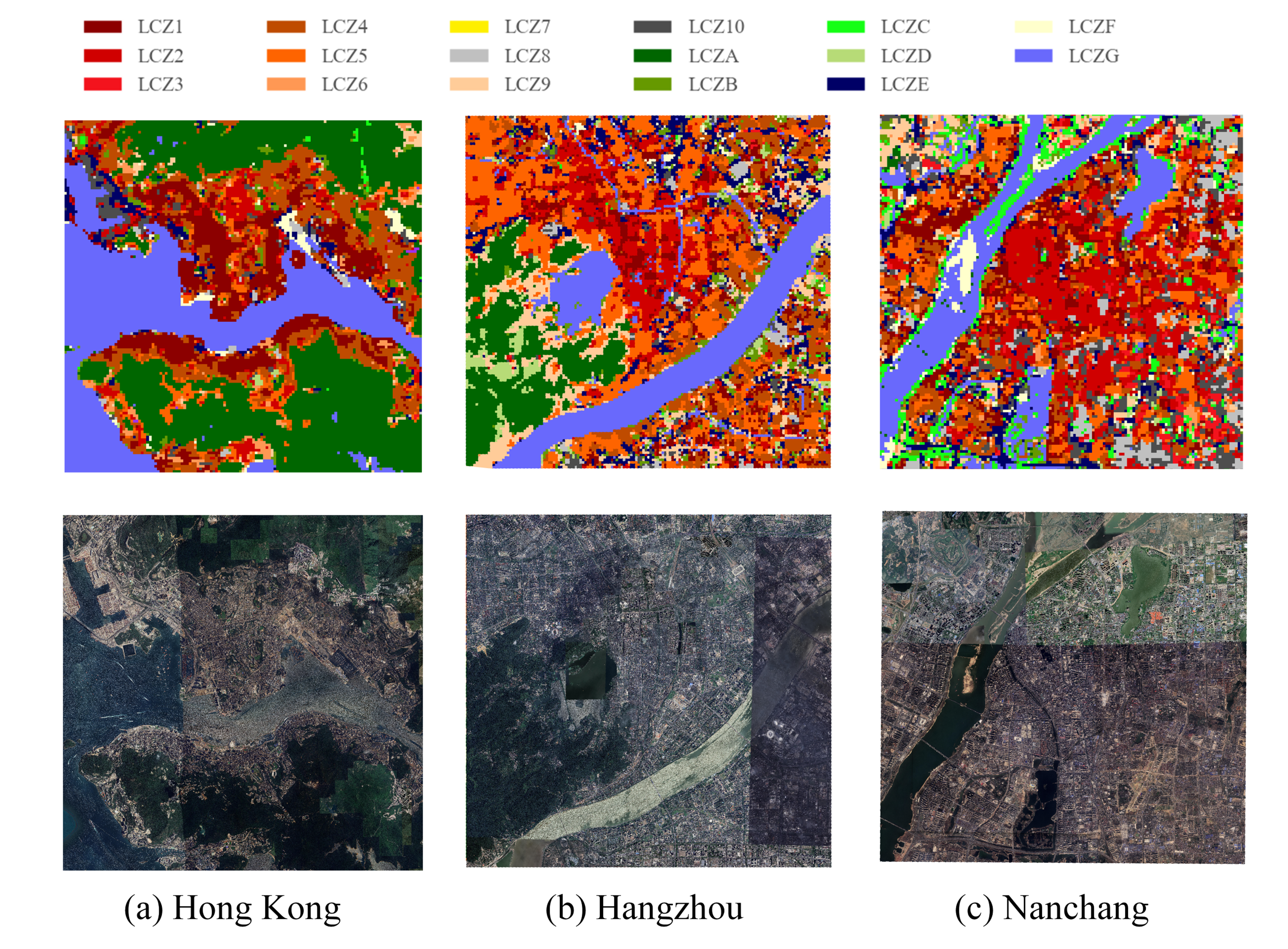}
	\caption{LCZ classifications of the three cities: Hong Kong, Hangzhou, and Nanchang. Top: LCZ maps; Bottom: Google Earth images.}
	\label{fig_exp1}
\end{figure*}

\subsection{Accuracy assessment}
Various metrics, derived from confusion matrices, are utilized to assess performance. These metrics encompass Overall Accuracy (OA), Overall Accuracy of Built-ups (OA$_{BU}$), Overall Accuracy of natural land covers (OA$_{N}$), Kappa coefficient, and Average F1 score. Overall Accuracy signifies the percentage of accurately classified samples across all classes. OA$_{BU}$ and OA$_{N}$ assess the model's accuracy in identifying built and natural land cover types within LCZ categories, respectively. Furthermore, the Kappa coefficient, denoted by $\kappa$, measures the agreement between predicted and actual classifications using the following formulas:
\begin{eqnarray}
	\kappa = \frac{P_o - P_e}{1 - P_e}			
\end{eqnarray}
where $P_o$ and $P_e$ are defined as:
\begin{eqnarray}
	P_{o}&=&\frac{\displaystyle\sum_{c=1}^{C} TP_{c}}{N}, \label{eq:Po}\\
	P_{e}&=&\frac{\displaystyle\sum_{c=1}^{C} (TP_c+FN_c)(TP_c+FP_c)}{N^{2}}.\label{eq:Pe}
\end{eqnarray}
with $TP_{c}$, $FP_{c}$, and $FN_{c}$ being true positives, false positives, and false negatives for the $c$-th class, respectively. Additionally, $N$ represents the total number of samples, while $C$ represents the total number of LCZ types.

Lastly, the Average F1 score computes the weighted average of the F1 scores for each class identified by the index $c$ using the following formulas:
\begin{eqnarray}
	\text{F1} = 2\times\frac{p_{c}r_{c}}{p_{c}+r_{c}},\label{eq8}
\end{eqnarray}
where $p_{c}$ and $r_{c}$ are given by:
\begin{eqnarray}
	p_{c}&=&\frac{TP_{c}}{TP_{c}+FP_{c}},\\\label{eq19}
	r_{c}&=&\frac{TP_{c}}{TP_{c}+FN_{c}}.\label{eq20}
\end{eqnarray}
Upon computing F1 for each class according to the definitions above, we derive their mean values, denoted as Average F1.

\begin{figure*}[h]
	\centering
	\includegraphics[width=0.95\linewidth]{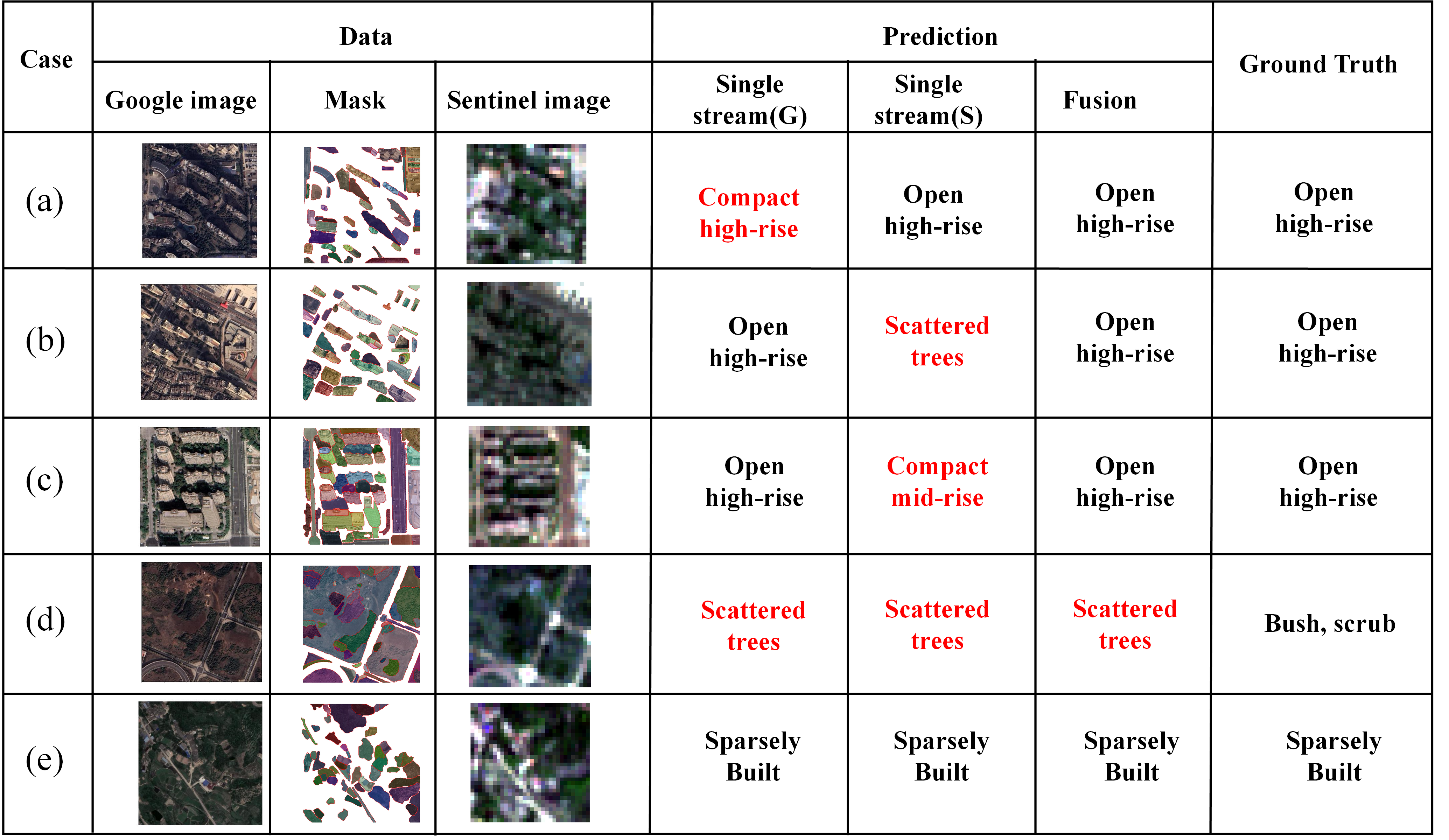}
	\caption{Case study examples. The data column presents the images	and ground instance masks used in the DF4LCZ model, while the prediction column provides corresponding results from single streams and the fusion in DF4LCZ. Additionally, the ground truth is shown in the last column. Classes highlighted in red represent misclassification results.}
	\label{fig_exp3}
\end{figure*}

\begin{figure*}[t]
	\centering
	\includegraphics[width=\linewidth]{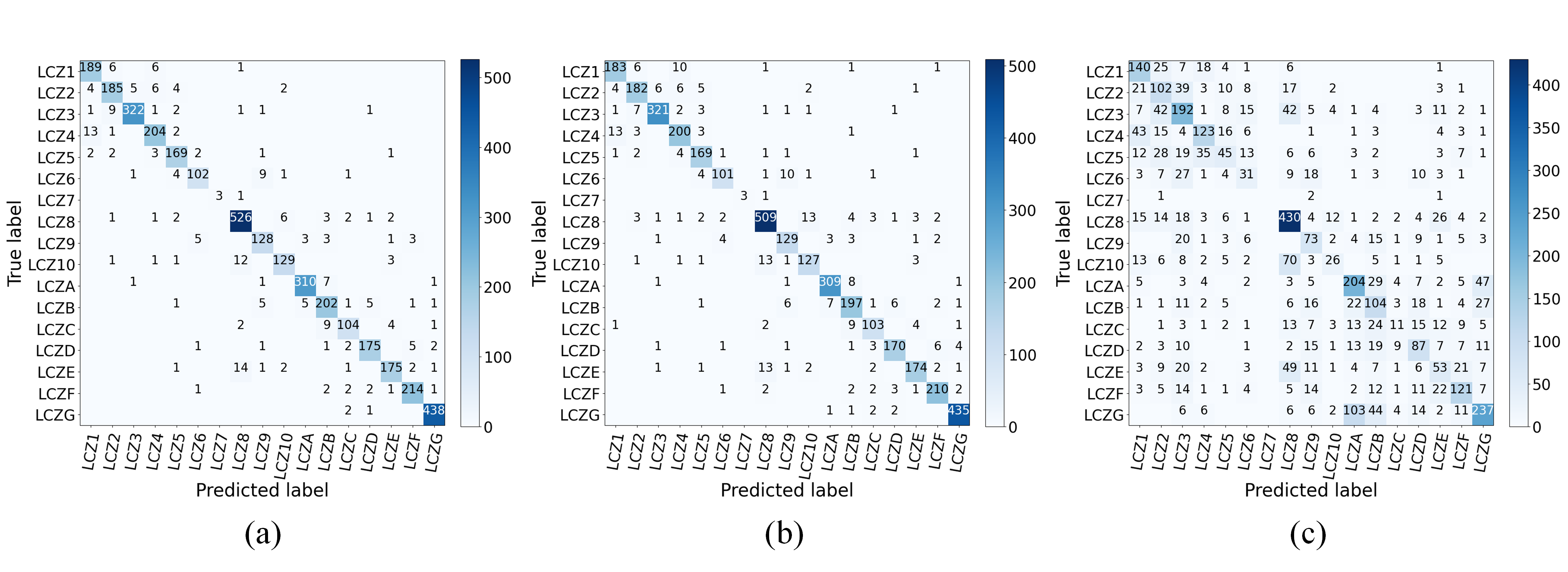}
	\caption{Confusion matrices using different data sources, including (a) DF4LCZ using both Sentinel imagery and Google images, (b) only Sentinel images, and (c) only Google images.}
	\label{fig_exp2}
\end{figure*}

\section{Results and discussion}\label{sec:results}
\subsection{LCZ classification and mapping results}
The overall classification results of the proposed dual-stream framework for scene-level local climate zone classification using Google Earth and Sentinel images are presented in Table~\ref{tab:results}.

As shown in Table~\ref{tab:results}, the proposed 3D ResNet11 branch achieved an OA of $92.42\%$ when using Sentinel-2 imagery alone, while the GCN-based branch achieved an OA of  $51.93\%$ when using Google images alone. However, when combining Sentinel-2 and Google Earth images (S+G), the classification accuracy was improved further, reaching an OA of $93.81\%$. Moreover, the scores of OA$_{BU}$ and OA$_{N}$ for the proposed dual-stream model reached $93.10\%$ and $94.68\%$, respectively, surpassing those achieved with single data sources alone. Furthermore, the fusion of Sentinel-2 and Google Earth images (S+G) achieved the highest Kappa coefficient values ($0.9329$), indicating substantial agreement between the predicted and actual classifications. Additionally, the Average F1 values ($93.80\%$) for the S+G case surpassed those of the single data source scenarios.

These results highlight the efficacy of the DF4LCZ framework in scene-level LCZ classification. The complementary nature of Google Earth and Sentinel-2 data enhances the robustness and discriminative capability of the classification model, resulting in improved performance metrics during testing.

Moreover, the DF4LCZ model is used for LCZ mapping of the cities. Fig.~\ref{fig_exp1} shows the mapping results for parts of Hong Kong, Hangzhou, and Nanchang, alongside corresponding reference Google Earth images. The results demonstrate that the mapping effectively classified highly dense and heterogeneous regions across these cities, aligning well with the reference images. Next, some case studies were examined, as depicted in Fig.~\ref{fig_exp3}. In this figure, the data column presents the images and ground instance masks used in the DF4LCZ model, while the prediction column provides corresponding results from single streams and the fusion in DF4LCZ. Additionally, the ground truth is shown in the last column. Classes highlighted in red represent misclassification results.

As shown in Fig.~\ref{fig_exp3}, the image patch is incorrectly classified as LCZ1 (compact high-rise) when utilizing a single stream of Google Earth images in Case (a). This misclassification might stem from incomplete and inaccurate ground instance extraction. However, Sentinel imagery provided additional spatial-spectral information, helping to avoid such misclassifications in the final fused result. In addition, some instances, such as those denoted by (b) and (c), exhibited incorrect predictions from the single stream of Sentinel imagery. Yet, due to complementary instance-based information from the Google image stream, correct predictions prevailed in the final fusion results. Furthermore, some instances, exemplified by (d), yielded erroneous predictions in both individual streams and the final fusion process. Such misclassifications may arise due to challenges in distinguishing height differences between trees and bushes using Google and Sentinel images. Finally, (e) illustrates an example of correct predictions.

\subsection{Per-class results and confusion matrices}
To evaluate DF4LCZ's ability to distinguish between different classes, Table~\ref{tab:perclass} presents the classification accuracy for each LCZ type. The classification accuracy for each LCZ type refers to the proportion of correctly classified samples within all samples classified as that type. The table highlights the variability in classification accuracy across various LCZ classes. For example, when solely utilizing Sentinel imagery, the accuracy ranges from 85.43\% to 100\%. However, combining the two data sources (S+G) results in higher accuracies across most LCZ classes compared to individual usage of each source. This indicates that our method effectively utilizes the advantages of both Google Earth and Sentinel imagery, leading to enhanced classification performance.


Further, three confusion matrices are shown in Fig.~\ref{fig_exp2}. The rows of the confusion matrices represent the true LCZ class labels, whereas the columns represent the predicted LCZ class labels. Every cell in the matrix indicates the count of test samples classified into a specific LCZ class. In the figure, it is clear that the numbers in the top right and bottom left cells in (a) are much smaller than those in the single data sources shown in (b) and (c), which implies that the dual-stream framework integrating both data sources can mitigate the misclassification between built types and land cover types within LCZs. Nevertheless, certain LCZ classes were still falsely classified, possibly stemming from the challenges of low inter-class variance and inadequate information for precise classification. This underscores the necessity of further comprehensive investigation in future studies.

\begin{table*}[t]
	\caption{Per-class testing accuracy of the DF4LCZ model.}
	\label{tab:perclass}
	\resizebox{\linewidth}{!}{
		\begin{tabular}{llllllllllllllllllll}
			\hline
			\multicolumn{1}{c}{\multirow{2}{*}{Data}} & \multicolumn{10}{c}{Built type}                                                                    &  & \multicolumn{7}{c}{Land cover}                                      & \multicolumn{1}{c}{\multirow{2}{*}{Overall}} \\ \cline{2-11} \cline{13-19}
			\multicolumn{1}{c}{}                      & 1       & 2       & 3       & 4       & 5       & 6       & 7        & 8       & 9       & 10      &  & A      & B      & C      & D      & E      & F      & G      & \multicolumn{1}{c}{}                         \\ \hline
			G                                         & 52.24\% & 39.38\% & 47.88\% & 60.59\% & 41.28\% & 32.98\% & 0.00\%   & 64.76\% & 39.25\% & 49.06\% &  & 54.84\% & 38.10\% & 29.73\% & 47.03\% & 33.76\% & 60.20\% & 67.91\% & 51.93\%                                      \\
			S                                         & 90.15\% & 89.22\% & 96.69\% & 89.29\% & 89.42\% & 91.82\% & 100.00\% & 93.57\% & 85.43\% & 86.99\% &  & 96.56\% & 86.78\% & 88.03\% & 92.90\% & 92.55\% & 93.33\% & 97.75\% & 92.42\%                                      \\
			S+G                                       & \textbf{90.43\%} & \textbf{90.24\%} & \textbf{97.87\%} & \textbf{91.89\%} & \textbf{90.86\%} & \textbf{91.89\%} & \textbf{100.00\%} & \textbf{94.43\%} & \textbf{87.07\%} & \textbf{92.14\%} &  & \textbf{97.48\%} & \textbf{88.99\%} & \textbf{90.43\%} & \textbf{94.59\%} & \textbf{93.58\%} & \textbf{95.11\%} & \textbf{98.43\%} & \textbf{93.81\%}                                      \\ \hline
	\end{tabular}}
\end{table*}

\subsection{Evaluation of different backbones}
In the proposed DF4LCZ model, we employ the 3D ResNet11 network as the backbone for processing Sentinel images. To evaluate the performance of this architecture on our dataset, we conducted evaluation experiments involving various backbone architectures. We substitute the 3D ResNet11 architecture with alternative backbone architectures, including MSMLA50~\cite{kim2021local}, ResNet50, ResNet11, DenseNet, and 3D ResNet11, in our DF4LCZ framework and compare their classification results. Additionally, we trained the selected backbone architectures as baselines for single-stream models to facilitate comparison with fusion models. Table~\ref{tab:results_3} presents the classification performance of the DF4LCZ model with different backbones and the corresponding single-stream baselines.

In the table, the DF4LCZ, which integrates data from Google Earth and Sentinel satellites, generally resulted in superior classification performance compared to models utilizing single data sources. The 3D ResNet11 backbone notably consistently outperformed alternative models across all evaluation metrics, indicating its efficacy in capturing spatial-spectral features.  

\begin{table*}[htbp]
	\centering
	\caption{Performance comparison between DF4LCZ and the baselines.}
	\label{tab:results_3}
	\begin{adjustbox}{width=0.7\textwidth}
		\begin{tabular}{lclllll}
			\hline
			\multicolumn{1}{c}{\multirow{2}{*}{Data}} & \multirow{2}{*}{Network} & \multicolumn{5}{c}{Test}                                                                                                           \\ \cline{3-7} 
			\multicolumn{1}{c}{}                      &                           & \multicolumn{1}{c}{OA} & \multicolumn{1}{c}{OA$_{BU}$} & \multicolumn{1}{c}{OA$_{N}$} & \multicolumn{1}{c}{Kappa} & \multicolumn{1}{c}{Avg. F1} \\ \hline
			S & MSMLA50 & 89.53\% & 89.44\% & 89.64\% & 0.8866 & 89.51\% \\
			& ResNet50 & 81.47\% & 81.16\% & 81.86\% & 0.7992 & 81.25\% \\
			& ResNet11 & 89.22\% & 88.49\% & 90.11\% & 0.8832 & 89.24\% \\
			& Densenet & 83.91\% & 83.87\% & 83.97\% & 0.8258 & 83.87\% \\
			& 3D ResNet11 & 92.42\% & 91.53\% & 93.50\% & 0.9179 & 92.42\% \\
			\hline
			S+G & DF4LCZ(MSMLA50) & 91.37\% & 91.48\% & 91.22\% & 0.9064 & 91.29\% \\
			& DF4LCZ(ResNet50) & 84.54\% & 84.40\% & 84.73\% & 0.8322 & 84.12\% \\
			& DF4LCZ(ResNet11) & 91.24\% & 91.06\% & 91.46\% & 0.9050 & 91.16\% \\
			& DF4LCZ(Densenet) & 86.70\% & 86.54\% & 86.89\% & 0.8558 & 86.50\% \\
			& DF4LCZ(3D ResNet11) & \textbf{93.81\%} & \textbf{93.10\%} & \textbf{94.68\%} & \textbf{0.9329} & \textbf{93.80\%} \\
			\hline\\
		\end{tabular}
	\end{adjustbox}
	\footnotesize \par
	Abbreviations: G - Google Earth Image, S - Sentinel Image
\end{table*}

\begin{table*}[]
	\centering
	\caption{Transferability assessment results of DF4LCZ and the baselines.}
	\label{tab:results_strategy}
	\begin{adjustbox}{width=0.7\textwidth}
		\begin{tabular}{lclllll}
			\hline
			\multicolumn{1}{c}{\multirow{2}{*}{Data}} & \multirow{2}{*}{Network} & \multicolumn{5}{c}{Test}                                                                                                           \\ \cline{3-7} 
			\multicolumn{1}{c}{}                      &                           & \multicolumn{1}{c}{OA} & \multicolumn{1}{c}{OA$_{BU}$} & \multicolumn{1}{c}{OA$_{N}$} & \multicolumn{1}{c}{Kappa} & \multicolumn{1}{c}{Avg. F1} \\ \hline
			S & MSMLA50 & 78.90\% & 80.69\% & 76.71\% & 0.7716 & 78.83\% \\
			& ResNet50 & 73.52\% & 73.26\% & 73.84\% & 0.7134 & 73.27\% \\
			& ResNet11 & 78.33\% & 77.64\% & 79.17\% & 0.7655 & 78.36\% \\
			& Densenet & 75.10\% & 75.40\% & 74.72\% & 0.7306 & 74.92\% \\
			& 3D ResNet11 & 80.74\% & 80.88\% & 80.57\% & 0.7916 & 80.72\% \\
			\hline
			S+G & DF4LCZ(MSMLA50) & 79.40\% & 81.26\% & 77.12\% & 0.7769 & 79.24\% \\
			& DF4LCZ(ResNet50) & 75.39\% & 75.40\% & 75.37\% & 0.7331 & 74.83\% \\
			& DF4LCZ(ResNet11) & 79.03\% & 79.31\% & 78.70\% & 0.7730 & 78.95\% \\
			& DF4LCZ(Densenet) & 76.31\% & 77.12\% & 75.31\% & 0.7433 & 75.90\% \\
			& DF4LCZ(3D ResNet11) & \textbf{81.13\%} & \textbf{81.35}\% &\textbf{ 80.87\%} &\textbf{0.7958} & \textbf{81.08\%} \\
			\hline\\
		\end{tabular}
	\end{adjustbox}
	\footnotesize \par
	Abbreviations: G - Google Earth Image, S - Sentinel Image
\end{table*}

\subsection{Transferability of the fusion models}

For all of the experiments described above, we employ the dataset sampled using the ``splitting the sample pool" strategy, as outlined in Section~\ref{sec:sampling}. In this subsection, we repeat the experiments using the dataset sampled through the ``splitting the polygon pool" strategy. This strategy ensures that fewer training and testing samples are drawn close to each other, thereby ensuring that testing samples are unseen during the training process. Consequently, we assess the transferability of our DF4LCZ model using this strategy. The overall results of these experiments are presented in Table~\ref{tab:results_strategy}.

Compared to the results presented in Table~\ref{tab:results_3}, the OAs of all models suffered from noticeable performance degradation. This decline can be attributed to a domain shift in the transferability test, where the models were evaluated using data from previously unseen regions. Such a shift implies that models trained on one domain may not generalize well when applied to a different domain, resulting in decreased accuracy. Furthermore, the Sentinel and Google Earth data (S+G) combination with the 3D ResNet11 structure consistently achieves the highest OA of $81.13\%$, demonstrating its robust performance across various sampling strategies. Concurrently, the fusion of Google Earth and Sentinel data typically yields enhanced OA and other evaluation metrics in comparison to the baselines utilizing individual data sources. These findings suggest that DF4LCZ exhibits superior transferability when tested on samples from previously unobserved regions.

\subsection{Impact of different weighted fusion parameters}

\begin{figure*}[htp]
	\centering
	\includegraphics[width=\linewidth]{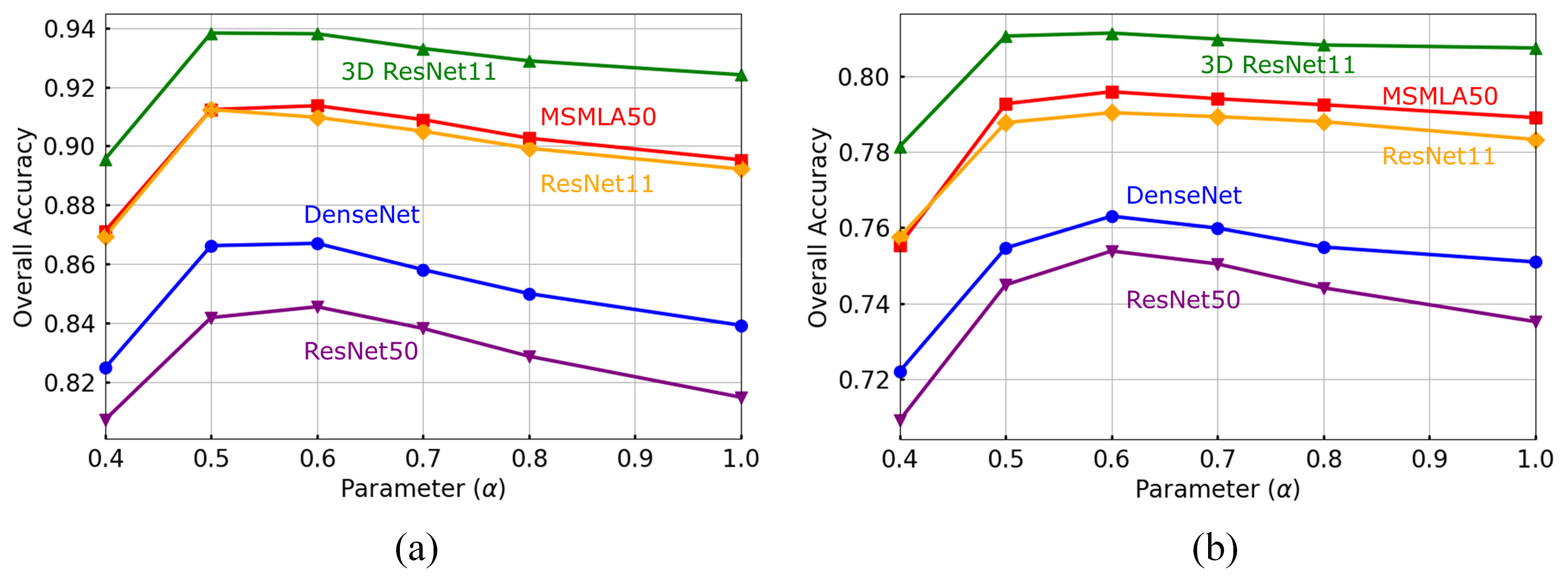}
	\caption{Performance comparison of the DF4LCZ framework using different backbones with varying weighted fusion parameters. Each colored line represents the OA variation using a specific backbone in the DF4LCZ framework. Subfigure (a) depicts the results obtained using the ``splitting the sample pool" strategy, while (b) presents the results obtained using the ``splitting the polygon pool" strategy.}
	\label{para}
\end{figure*}

In this section, we analyze the impact of the weighting coefficient  $\alpha$ on the performance of DF4LCZ. Experiments were performed on multi-source datasets using both splitting strategies discussed. Additionally, DF4LCZ utilizes different backbones for the Sentinel image stream. Fig.~\ref{para} illustrates DF4LCZ's performance as a function of $\alpha$.

In the figure, irrespective of the dataset or backbone architecture used, all lines exhibit similar patterns. For each line, as the parameter $\alpha$ increased, the OA initially rose and then peaked around $0.5$ or $0.6$. These findings suggest that adjusting the fusion parameter $\alpha$ allows for fine-tuning the balance between Google Earth and Sentinel image contributions, thus maximizing classification accuracy. In practical applications, practitioners can choose the $\alpha$ value that maximizes the desired performance metric(s) for their specific needs.

\section{Conclusion}\label{sec:conclusion}
This work has proposed DF4LCZ, a SAM-empowered data fusion framework tailored to enhance the accuracy of local climate zone classification by combining high-resolution Google RGB images with the multispectral imagery from Sentinel-2. Unlike traditional single-stream CNN-based models, DF4LCZ adeptly combines instance-based location features from Google imagery and scene-level spatial-spectral features from Sentinel-2 imagery. In particular, a SAM-assisted GCN branch has been developed to extract the high-level instance-based location features from the Google Earth imagery, and a 3D ResNet11 branch is designed to extract the scene-level spatial-spectral features from Sentinel-2 imagery. A new LCZ dataset called LCZC-GES2, composed of both Google and Sentinel-2 imagery across eight cities in Southeast China, has been developed. Extensive computer simulation on the newly developed dataset LCZC-GES2 has confirmed that the proposed dual-stream framework can substantially outperform conventional single-stream LCZ classification models. 

\small
\bibliographystyle{IEEEtran}
\bibliography{refs}
\end{document}